\def\year{2019}\relax
\documentclass[letterpaper]{article} %
\usepackage{aaai19}  %
\usepackage{times}  %
\usepackage{helvet}  %
\usepackage{courier}  %
\usepackage{url}  %
\usepackage{graphicx}  %
\graphicspath{ {images/} }
\usepackage{multirow}
\usepackage{amsmath}
\usepackage{amssymb}
\usepackage{xcolor}
\usepackage{cleveref}

\DeclareMathOperator\myrelu{\operatorname{ReLU}}
\DeclareMathOperator\sigmoid{\operatorname{sigmoid}}
\DeclareMathOperator\avg{\operatorname{avg}}
\DeclareMathOperator\softmax{\operatorname{softmax}}
\DeclareMathOperator*{\argmax}{arg\,max}
\newcommand{\lform}[1]{\textsf{\fontsize{7.5pt}{.1pt}\selectfont{#1}}}
\makeatletter
\newcommand{\thickhline}{%
    \noalign {\ifnum 0=`}\fi \hrule height 1pt
    \futurelet \reserved@a \@xhline
}
\makeatother
\nocopyright
\begin{document}
\title{Data-to-Text Generation with Content Selection and Planning}
 \author{Ratish Puduppully  \textnormal{and} Li Dong \textnormal{and} Mirella Lapata\\
Institute for Language, Cognition and Computation\\
School of Informatics, University of Edinburgh\\
 10 Crichton Street, Edinburgh EH8 9AB\\
\texttt{r.puduppully@sms.ed.ac.uk}~~~~\texttt{li.dong@ed.ac.uk}~~~~\texttt{mlap@inf.ed.ac.uk}\\
}
 
\maketitle
\begin{abstract}

  Recent advances in data-to-text generation have led to the use of
  large-scale datasets and neural network models which are trained
  end-to-end, without explicitly modeling \emph{what to say} and
  \emph{in what order}. In this work, we present a neural network
  architecture which incorporates content selection and planning
  without sacrificing end-to-end training. We decompose the generation
  task into two stages. Given a corpus of data records (paired with
  descriptive documents), we first generate a content plan
  highlighting which information should be mentioned and in which
  order and then generate the document while taking the content plan
  into account. Automatic and human-based evaluation experiments show
  that our model\footnote{Our code is publicly available at
    \url{https://github.com/ratishsp/data2text-plan-py}.} outperforms strong baselines improving the
  state-of-the-art on the recently released \textsc{RotoWire} dataset.

\end{abstract}
\section{Introduction}
Data-to-text generation broadly refers to the task of automatically
producing text from non-linguistic input
\cite{reiter-dale:00,gatt2018survey}. The input may be in various forms
including databases of records, spreadsheets, expert system knowledge bases,
simulations of physical systems, and so on. Table~\ref{fig:example}
shows an example in the form of a database containing statistics on
NBA basketball games, and a corresponding game summary.

\begin{table*}
\begin{small}
\begin{minipage}[c]{0.37\linewidth}

\begin{tabular}{@{}l@{~~}r@{~~}c@{~~}r@{~~}c@{~~}c@{~~}c@{~~}l@{}} \hline
\lform{TEAM}      & \lform{WIN} &\lform{LOSS} &\lform{PTS} & \lform{FG\_PCT} & \lform{RB} & \lform{AST} & \lform{$\dots$} \\ \hline
\lform{Pacers} &\lform{4} &\lform{6} &\lform{99} & \lform{42} & \lform{40} & \lform{17}& \lform{$\dots$} \\ 
\lform{Celtics} &\lform{5} &\lform{4} &\lform{105} & \lform{44} &\lform{47} &\lform{22} &\lform{$\dots$} \\ \hline
\end{tabular}

\begin{tabular}{@{}l@{~~}c@{~~}c@{~}r@{~~}c@{~~}c@{~~}l@{~~}l@{}} \hline
\lform{PLAYER} & \lform{H/V} & \lform{AST} & \lform{RB} & \lform{PTS} & \lform{FG} & \lform{CITY} & \lform{$\dots$}\\ \hline
\lform{Jeff Teague} & \lform{H} & \lform{4} & \lform{3} & \lform{20} & \lform{4} & \lform{Indiana} & \lform{$\dots$}\\
\lform{Miles Turner} & \lform{H} & \lform{1} & \lform{8} & \lform{17} & \lform{6} & \lform{Indiana} & \lform{$\dots$}\\
\lform{Isaiah Thomas} & \lform{V}& \lform{5} & \lform{0} & \lform{23} & \lform{4} & \lform{Boston} & \lform{$\dots$}\\
\lform{Kelly Olynyk} & \lform{V}& \lform{4} & \lform{6} & \lform{16} & \lform{6} & \lform{Boston} & \lform{$\dots$}\\
\lform{Amir Johnson} & \lform{V}& \lform{3} & \lform{9} & \lform{14} & \lform{4} & \lform{Boston} & \lform{$\dots$}\\
\lform{$\dots$} & \lform{$\dots$} & \lform{$\dots$} & \lform{$\dots$} & \lform{$\dots$}& \lform{$\dots$} & \lform{$\dots$}\\\hline

\end{tabular}\\
\lform{PTS:} points, \lform{FT\_PCT:} free throw percentage,
\lform{RB:}  rebounds, \lform{AST:} 
assists, \lform{H/V:} home or visiting, \lform{FG:} field goals,
\lform{CITY:} player team city.
\end{minipage}
\begin{minipage}[c]{0.1\linewidth}
  \begin{tabular}{p{10.7cm}} \hline The \textbf{Boston Celtics}
    defeated the host \textbf{Indiana Pacers} \textbf{105-99} at
    Bankers Life Fieldhouse on Saturday. In a battle between two
    injury-riddled teams, the Celtics were able to prevail with a much
    needed road victory. The key was shooting and defense, as the
    \textbf{Celtics} outshot the \textbf{Pacers} from the field, from
    three-point range and from the free-throw line. Boston also held
    Indiana to \textbf{42 percent} from the field and \textbf{22
      percent} from long distance. The Celtics also won the rebounding
    and assisting differentials, while tying the Pacers in
    turnovers. There were 10 ties and 10 lead changes, as this game
    went down to the final seconds. Boston (\textbf{5--4}) has had to deal with
    a gluttony of injuries, but they had the fortunate task of playing
    a team just as injured here. \textbf{Isaiah} Thomas led the team in
    scoring, totaling \textbf{23 points and five assists on 4--of--13}
    shooting. He got most of those points by going 14--of--15 from the
    free-throw line. \textbf{Kelly Olynyk} got a rare start and finished second
    on the team with his \textbf{16 points, six rebounds and four assists}.
\end{tabular}
\end{minipage}
\end{small}
\label{fig:example}
\caption{Example of data-records and document
  summary. Entities and values corresponding to the plan in
  Table~\ref{tbl:example-content-plan} are boldfaced.}
\end{table*}

Traditional methods for data-to-text generation \cite{P83-1022,mckeown1992text} implement a pipeline
of modules including content planning (selecting specific content from some input and determining the
structure of the output text), sentence planning (determining the structure and lexical content of each sentence)
 and surface realization (converting the sentence plan to a surface string).
Recent neural generation systems (Lebret et
al.~\citeyear{D16-1128}; Mei et al.~\citeyear{N16-1086}; Wiseman et
al.~\citeyear{wiseman2017challenges}) do not explicitly model any of
these stages, rather they are trained in an end-to-end fashion using
the very successful encoder-decoder architecture (Bahdanau et
al.~\citeyear{bahdanau2015neural}) as their backbone.

\begin{figure}[t]
\includegraphics[scale=0.7]{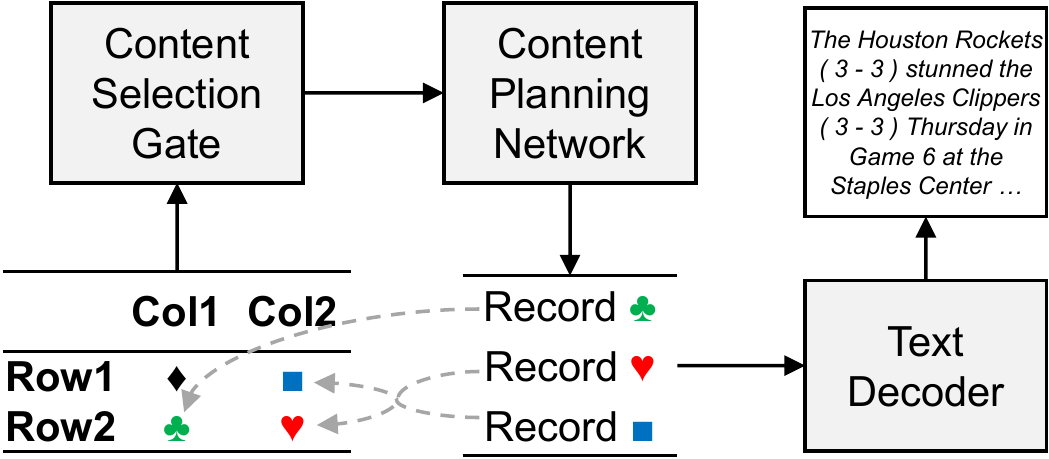}
\caption{Block diagram of our approach.}
\label{fig:rotowire-example}
\end{figure}

Despite producing overall fluent text, neural systems have difficulty
capturing long-term structure and generating documents more than a few
sentences long. Wiseman et al.~\shortcite{wiseman2017challenges} show
that neural text generation techniques perform poorly at content
selection, they struggle to maintain inter-sentential coherence, and
more generally a reasonable ordering of the selected facts in the
output text. Additional challenges include avoiding redundancy and
being faithful to the input. Interestingly, comparisons against
template-based methods show that neural techniques do not fare well on
metrics of content selection recall and factual output generation
(i.e.,~they often hallucinate statements which are not supported by
the facts in the database).

In this paper, we address these shortcomings by \emph{explicitly}
modeling content selection and planning within a neural data-to-text
architecture.  Our model learns a content plan from the input and
conditions on the content plan in order to generate the output
document (see Figure~\ref{fig:example} for an illustration). An
explicit content planning mechanism has at least three advantages for
multi-sentence document generation: it represents a high-level
organization of the document structure allowing the decoder to
concentrate on the easier tasks of sentence planning and surface
realization; it makes the process of data-to-document generation more
interpretable by generating an intermediate representation; and
reduces redundancy in the output, since it is less likely for the
content plan to contain the same information in multiple places.

We train our model end-to-end using neural networks and evaluate
its performance on \textsc{RotoWire} (Wiseman et
al.~\citeyear{wiseman2017challenges}), a recently released dataset
which contains statistics of NBA basketball games paired with
human-written summaries (see Table~\ref{fig:example}). Automatic and
human evaluation shows that modeling content selection and
planning improves generation considerably over competitive baselines.

\section{Related Work}

The generation literature provides multiple examples of content
selection components developed for various domains which are either
hand-built
\cite{P83-1022,mckeown1992text,Reiter:1997:BAN:974487.974490,W03-1016}
or learned from data (Barzilay and Lapata~\citeyear{H05-1042}; Duboue
and McKeown~\citeyear{duboue2001empirically,W03-1016}; Liang et
al.~\citeyear{P09-1011}; Angeli et al.~\citeyear{D10-1049}; Kim and
Mooney~\citeyear{C10-2062}; Konstas and
Lapata \citeyear{Konstas:2013:GMC:2591248.2591256}). Likewise,
creating summaries of sports games has been a topic of interest since
the early beginnings of generation systems
(Robin~\citeyear{robin1994revision}; Tanaka-Ishii et
al.~\citeyear{P98-2209}).

Earlier work on content planning has relied on generic planners \cite{dale1988generating},
based on Rhetorical Structure Theory \cite{hovy1993automated} and schemas \cite{mckeown1997language}.
They defined content planners by analysing the target texts and devising hand-crafted rules. 
Duboue and McKeown \shortcite{duboue2001empirically}
studied ordering constraints for content plans and Duboue and McKeown \shortcite{duboue2002content} learn a content planner 
from an aligned corpus of inputs and human outputs. 
A few researchers \cite{mellish1998experiments,karamanis2004entity} rank 
content plans according to a ranking function.

More recent work focuses on end-to-end systems instead of individual components.
However, most models make simplistic assumptions such as generation
without any content selection or planning \cite{Belz:2008:AGW:1520025.1520026,N07-1022} 
or content selection without planning  (Konstas and
Lapata~\citeyear{N12-1093}; Angeli et al.~\citeyear{D10-1049}; Kim and
Mooney \citeyear{C10-2062}).
An exception are Konstas and Lapata
\shortcite{Konstas:2013:GMC:2591248.2591256} who incorporate content
plans represented as grammar rules operating on the document
level. Their approach works reasonably well with weather forecasts,
but does not scale easily to larger databases, with richer vocabularies, and
longer text descriptions. The model relies on the EM algorithm
\cite{dempster1977maximum} to learn the weights of the grammar rules
which can be very many even when tokens are aligned to database
records as a preprocessing step.

Our work is closest to recent neural network models which learn
generators from data and accompanying text resources. Most previous
approaches generate from Wikipedia infoboxes focusing either on single
sentences (Lebret et al.~\citeyear{D16-1128,E17-1060}; Sha et
al.~\citeyear{sha2017order}; Liu et al.~\citeyear{liu2017table}) or
short texts \cite{perez2018bootstrapping}.  Mei et
al.~\shortcite{N16-1086} use a neural encoder-decoder model to
generate weather forecasts and soccer commentaries, while Wiseman et
al.~\shortcite{wiseman2017challenges} generate NBA game summaries (see
Table~\ref{fig:rotowire-example}). They introduce a new dataset for
data-to-document generation which is sufficiently large for neural
network training and adequately challenging for testing the
capabilities of document-scale text generation (e.g., the average
summary length is~330 words and the average number of input records
is~628). Moreover, they propose various automatic evaluation measures
for assessing the quality of system output. Our model follows on from
Wiseman et al.~\shortcite{wiseman2017challenges} addressing the
challenges for data-to-text generation identified in their work.  We
are not aware of any previous neural network-based approaches which
incorporate content selection and planning mechanisms and generate
multi-sentence documents. \citeauthor{perez2018bootstrapping}
\shortcite{perez2018bootstrapping} introduce a content selection
component (based on multi-instance learning) without content planning,
while \citeauthor{liu2017table} \shortcite{liu2017table} propose a
sentence planning mechanism which orders the contents of a Wikipedia
infobox in order to generate a single sentence.

\section{Problem Formulation}

\begin{figure}[t]
\begin{center}
\includegraphics[scale=.42]{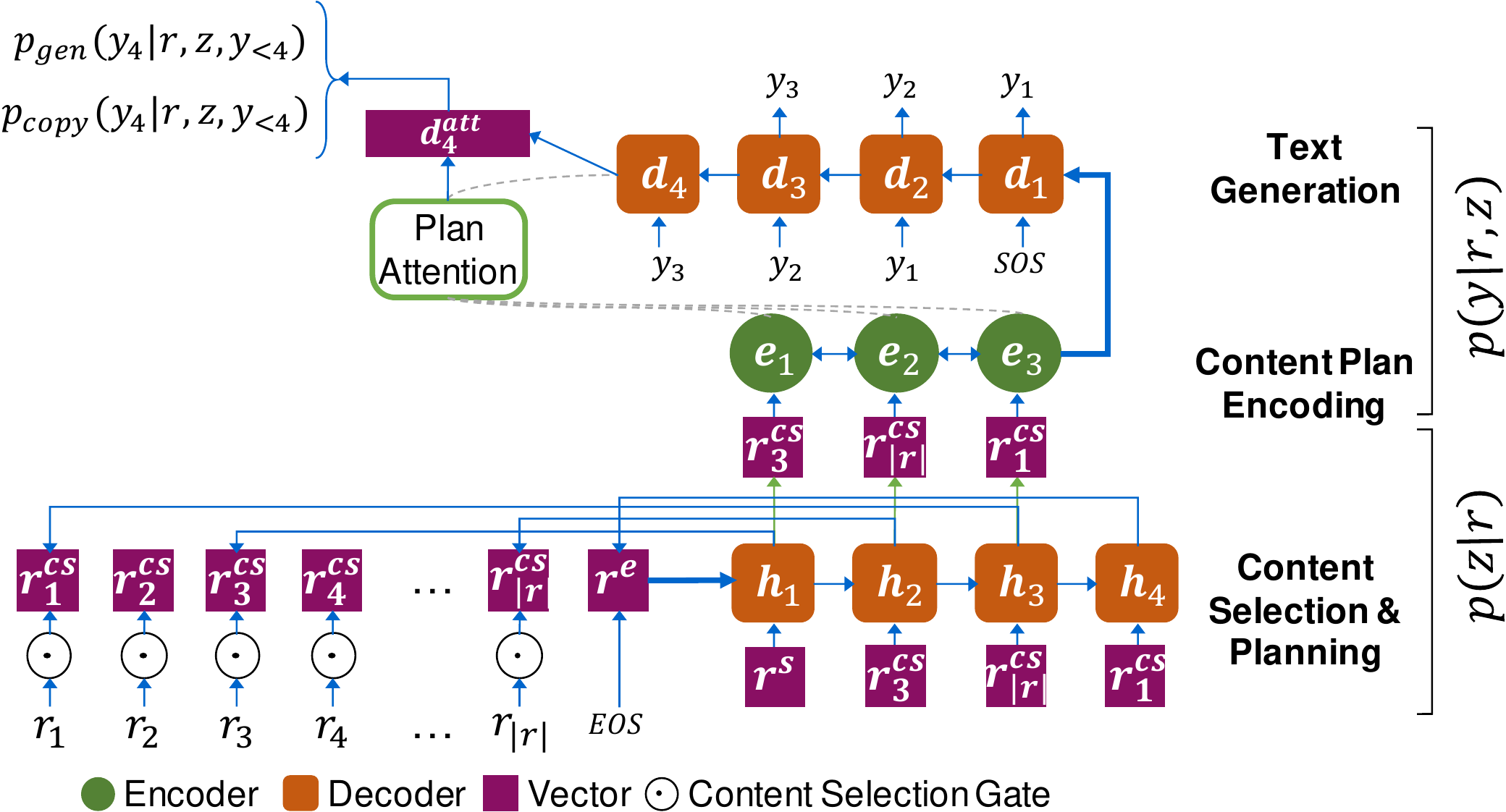}
\end{center}
\caption{Generation model with content selection and planning;
  the content selection gate is illustrated in
  Figure~\ref{fig:content-selection}.} 
\label{fig:overall-method}
\end{figure}

The input to our model is a table of records (see
Table~\ref{fig:example} left hand-side). Each record~$r_j$ has four
features including its type ($r_{j,1}$; e.g.,~\lform{\small LOSS},
\lform{\small CITY}), entity ($r_{j,2}$; e.g.,~\lform{\small Pacers},
\lform{\small Miles Turner}), value ($r_{j,3}$; e.g.,~\lform{\small
  11}, \lform{\small Indiana}), and whether a player is on the home-
or away-team ($r_{j,4}$; see column~\lform{\small H/V} in
Table~\ref{fig:example}), represented as \{$r_{j,k}\}_{k=1}^{4}$. The
output $y$ is a document containing words $y = y_1 \cdots y_{|y|}$
where $|y|$~is the document length.  Figure~\ref{fig:overall-method}
shows the overall architecture of our model which consists of two
stages: (a) \emph{content selection and planning} operates on the input
records of a database and produces a content plan specifying which
records are to be verbalized in the document and in which order (see
Table~\ref{tbl:example-content-plan}) and (b)~\emph{text generation}
produces the output text given the content plan as input; at each
decoding step, the generation model attends over vector
representations of the records in the content plan.

Let~$r=\{r_j\}_{j=1}^{|r|}$ denote a table of input records and~$y$
the output text. We model~$p(y|r)$ as the joint probability of
text~$y$ and content plan~$z$, given input~$r$. We further
decompose~$p(y,z|r)$ into~$p(z|r)$, a content selection and planning
phase, and $p(y|r,z)$, a text generation phase:
\begin{equation}
p(y|r) = \sum_z p(y,z|r) = \sum_z p(z|r)p(y|r,z) \nonumber
\end{equation}
In the following we explain how the components~$p(z|r)$ and $p(y|r,z)$
are estimated.

\subsection{Record Encoder}
\label{sec:record-encoder}

The input to our model is a table of unordered records, each
represented as features~\{$r_{j,k}\}_{k=1}^{4}$. Following previous
work (Yang et al.~\citeyear{D17-1197}; Wiseman et
al.~\citeyear{wiseman2017challenges}), we embed features into vectors,
and then use a multilayer perceptron to obtain a vector representation
$\mathbf{r}_j$ for each record:
\begin{equation}
\mathbf{r}_j = \myrelu(\mathbf{W}_r[\mathbf{r}_{j,1};\mathbf{r}_{j,2};\mathbf{r}_{j,3};\mathbf{r}_{j,4}]+\mathbf{b}_r) \nonumber
\end{equation}
where $[;]$ indicates vector concatenation, $\mathbf{W}_r \in \mathbb{R}^{n \times 4n} , \mathbf{b}_r \in \mathbb{R}^{n}$ are parameters, and $\myrelu$ is the rectifier activation function.

\subsection{Content Selection Gate}
\label{content-selection-gate}
\begin{figure}[t]
\centering
\includegraphics[width=\linewidth]{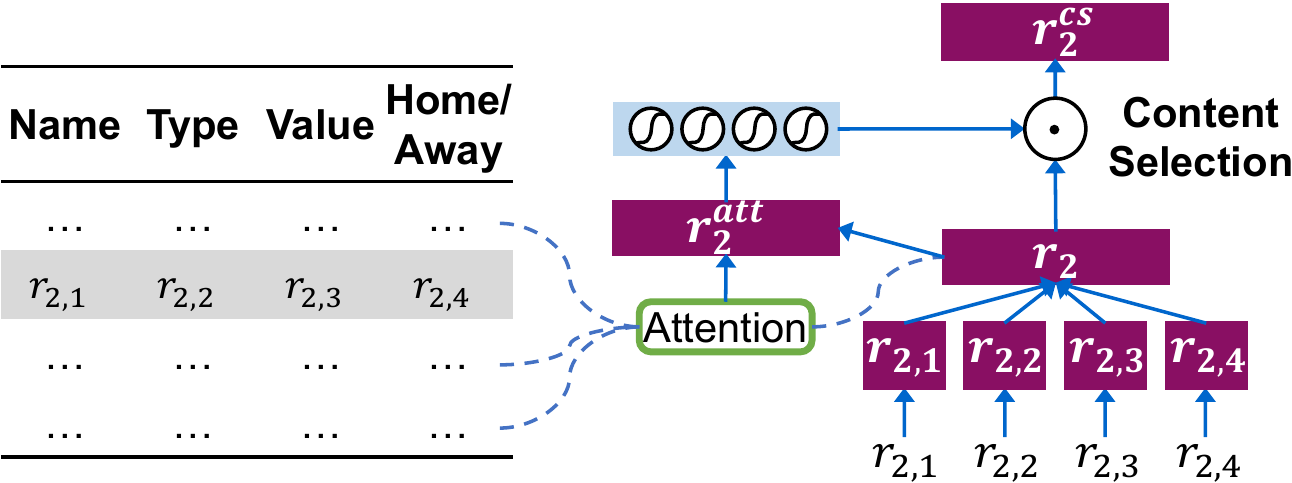}
\caption{Content selection mechanism.
\label{fig:content-selection}}
\end{figure}

The context of a record can be useful in determining its importance
vis-a-vis other records in the table. For example, if a player scores
many points, it is likely that other meaningfully related records such
as field goals, three-pointers, or rebounds will be mentioned in the
output summary. To better capture such dependencies among records, we
make use of the content selection gate mechanism as shown in
Figure~\ref{fig:content-selection}.

We first compute the attention scores~$\alpha_{j,k}$ over the
input table and use them to obtain an attentional vector $\mathbf{r}_j^{att}$~for each record~$r_j$:
\begin{align}
\alpha_{j,k} &\propto \exp (\mathbf{r}_j^\intercal \mathbf{W}_a \mathbf{r}_k) \nonumber \\
\mathbf{c}_j &= \sum_{k \neq j} \alpha_{j,k}\mathbf{r}_{k} \nonumber \\
\mathbf{r}_j^{att} &= \mathbf{W}_g [\mathbf{r}_j; \mathbf{c}_j] \nonumber
\end{align}
where $\mathbf{W}_a \in \mathbb{R}^{n \times n} , \mathbf{W}_g \in \mathbb{R}^{n \times 2n}$ are parameter matrices, and $\sum_{k \neq j} \alpha_{j,k} = 1$.

We next apply the content selection gating mechanism
to~$\mathbf{r}_j$, and obtain the new record
representation~$\mathbf{r}_j^{cs}$ via:
\begin{align}
\mathbf{g}_j &= \sigmoid \left( \mathbf{r}_j^{att} \right) \nonumber \\
\mathbf{r}_j^{cs} &= \mathbf{g}_j \odot \mathbf{r}_j \nonumber
\end{align}
where $\odot$ denotes element-wise multiplication, and gate~$\mathbf{g}_j
\in {[0,1]}^{n}$ controls the amount of information flowing
from~$\mathbf{r}_j$.  In other words, each element in~$\mathbf{r_j}$
is weighed by the corresponding element of the content selection gate
$\mathbf{g}_j$.%

\subsection{Content Planning}
In our generation task, the output text is long but follows a
canonical structure.  Game summaries typically begin by discussing
which team won/lost, following with various statistics involving
individual players and their teams (e.g., who performed exceptionally
well or under-performed), and finishing with any upcoming games.  We
hypothesize that generation would benefit from an explicit plan
specifying both {\it what to say} and {\it in which order}. Our model
learns such content plans from training data.  However, notice that
\textsc{RotoWire} (see Table~\ref{fig:example}) and most similar
data-to-text datasets do not naturally contain content plans.
Fortunately, we can obtain these relatively straightforwardly
following an information extraction approach (which we explain in
Section~\ref{sec:experimental-setup}).

Suffice it to say that plans are extracted by mapping the text in the
summaries onto entities in the input table, their values, and types
(i.e.,~relations). A plan is a sequence of pointers with each entry
pointing to an input record $\{ r_j \}_{j=1}^{|r|}$.  An excerpt of a
plan is shown in Table~\ref{tbl:example-content-plan}. The order in
the plan corresponds to the sequence in which entities appear in the
game summary. Let $z = z_1 \dots z_{|z|}$ denote the content planning
sequence.  Each $z_k$ points to an input record, i.e.,~$z_k \in \{ r_j
\}_{j=1}^{|r|}$.  Given the input records, the probability $p(z|r)$ is
decomposed as:
\begin{equation}
p(z|r) = \prod_{k=1}^{|z|} {p(z_k | z_{<k},r)} \nonumber
\end{equation}
where $z_{<k} = z_1 \dots z_{k-1}$.

Since the output tokens of the content planning stage correspond to
positions in the input sequence, we make use of Pointer Networks
(Vinyals et al.~\citeyear{NIPS2015_5866}). The latter use attention to
point to the tokens of the input sequence rather than creating a
weighted representation of source encodings. As shown in
Figure~\ref{fig:overall-method}, given $\{ r_j \}_{j=1}^{|r|}$, we use
an LSTM decoder to generate tokens corresponding to positions in the
input.  The first hidden state of the decoder is initialized by $\avg(
\{ \mathbf{r}_j^{cs} \}_{j=1}^{|r|} )$, i.e., the average of record
vectors.  At decoding step~$k$, let $\mathbf{h}_k$ be the hidden state
of the LSTM.  We model $p(z_k=r_j |z_{<k},r)$ as the attention over
input records:
\begin{align}
p(z_k = r_j | z_{<k} , r) \propto \exp (\mathbf{h}_k^\intercal \mathbf{W}_c \mathbf{r}_j^{cs}) \nonumber
\end{align}
where the probability is normalized to $1$, and $\mathbf{W}_c$ are
parameters.  Once $z_k$ points to record~$r_j$, we use the
corresponding vector $\mathbf{r}_j^{cs}$ as the input of the next LSTM
unit in the decoder.

\begin{table}[t]
\small
\begin{center}
\begin{tabular}{|l|l|l|l|}
\hline 
\multicolumn{1}{|c|}{\lform{Value}}&\multicolumn{1}{c|}{\lform{Entity}}&\multicolumn{1}{c|}{\lform{Type}}&\multicolumn{1}{c|}{\lform{H/V}}\\
\hline 
\lform{Boston}&\lform{Celtics}&\lform{TEAM-CITY}&\lform{V}\\
\lform{Celtics}&\lform{Celtics}&\lform{TEAM-NAME}&\lform{V}\\
\lform{105}&\lform{Celtics}&\lform{TEAM-PTS}&\lform{V}\\
\lform{Indiana}&\lform{Pacers}&\lform{TEAM-CITY}&\lform{H}\\
\lform{Pacers}&\lform{Pacers}&\lform{TEAM-NAME}&\lform{H}\\
\lform{99}&\lform{Pacers}&\lform{TEAM-PTS}&\lform{H}\\
\lform{42}&\lform{Pacers}&\lform{TEAM-FG\_PCT}&\lform{H}\\
\lform{22}&\lform{Pacers}&\lform{TEAM-FG3\_PCT}&\lform{H}\\
\lform{5}&\lform{Celtics}&\lform{TEAM-WIN}&\lform{V}\\
\lform{4}&\lform{Celtics}&\lform{TEAM-LOSS}&\lform{V}\\
\lform{Isaiah}&\lform{Isaiah\_Thomas}&\lform{FIRST\_NAME}&\lform{V}\\
\lform{Thomas}&\lform{Isaiah\_Thomas}&\lform{SECOND\_NAME}&\lform{V}\\
\lform{23}&\lform{Isaiah\_Thomas}&\lform{PTS}&\lform{V}\\
\lform{5}&\lform{Isaiah\_Thomas}&\lform{AST}&\lform{V}\\
\lform{4}&\lform{Isaiah\_Thomas}&\lform{FGM}&\lform{V}\\
\lform{13}&\lform{Isaiah\_Thomas}&\lform{FGA}&\lform{V}\\
\lform{Kelly}&\lform{Kelly\_Olynyk}&\lform{FIRST\_NAME}&\lform{V}\\
\lform{Olynyk}&\lform{Kelly\_Olynyk}&\lform{SECOND\_NAME}&\lform{V}\\
\lform{16}&\lform{Kelly\_Olynyk}&\lform{PTS}&\lform{V}\\
\lform{6}&\lform{Kelly\_Olynyk}&\lform{REB}&\lform{V}\\
\lform{4}&\lform{Kelly\_Olynyk}&\lform{AST}&\lform{V}\\
$\dots$ & $\dots$ & $\dots$ & $\dots$ \\ \hline
\end{tabular} 
\end{center}
\caption{Content plan for the example in Table~\ref{fig:example}.}
\label{tbl:example-content-plan}

\end{table}

\subsection{Text Generation}
The probability of output text~$y$ conditioned on content plan~$z$ and
input table~$r$ is modeled as:
\begin{equation}
p(y|r,z) = \prod_{t=1}^{|y|} p(y_t|y_{<t},z,r) \nonumber
\end{equation}
where $y_{<t} = y_1 \dots y_{t-1}$.
We use the encoder-decoder architecture with an attention mechanism to compute $p(y|r,z)$.

We first encode the content plan $z$ into
$\{ \mathbf{e}_k \}_{k=1}^{|z|}$ using a bidirectional LSTM. Because
the content plan is a sequence of input records, we directly feed the
corresponding record vectors $\{ \mathbf{r}_j^{cs} \}_{j=1}^{|r|}$ as
input to the LSTM units, which share the record encoder with the first
stage.

The text decoder is also based on a recurrent neural network with LSTM units.
The decoder is initialized with the hidden states of the final step in the encoder.
At decoding step $t$, the input of the LSTM unit is the embedding of the previously predicted word $y_{t-1}$.
Let $\mathbf{d}_t$ be the hidden state of the $t$-th LSTM unit.
The probability of predicting $y_t$ from the output vocabulary is computed via:
\begin{align}
\beta_{t,k} &\propto \exp (\mathbf{d}_t^\intercal \mathbf{W}_b \mathbf{e}_k) \label{eq:beta:attention} \\
\mathbf{q}_t &= \sum_k \beta_{t,k} \mathbf{e}_k \nonumber \\
\mathbf{d}_t^{att} &= \tanh( \mathbf{W}_d [ \mathbf{d}_t ; \mathbf{q}_t ] ) \nonumber \\
\hspace*{-.7cm}p_{gen} (y_t | y_{<t},z,r) &\hspace*{-.5ex}=\hspace*{-.5ex}\softmax_{y_t}{\hspace*{-.5ex}( \mathbf{W}_y \mathbf{d}_t^{att} + \mathbf{b}_y )}\hspace*{-2ex} \label{eq:prb:gen}
\end{align}
where $\sum_{k} \beta_{t,k} = 1$, $\mathbf{W}_b \in \mathbb{R}^{n \times n} , \mathbf{W}_d \in \mathbb{R}^{n \times 2n} , \mathbf{W}_y \in \mathbb{R}^{n \times |\mathcal{V}_y|} , \mathbf{b}_y \in \mathbb{R}^{|\mathcal{V}_y|}$ are parameters, and $|\mathcal{V}_y|$ is the output vocabulary size.

We further augment the decoder with a copy mechanism, i.e., the ability to
copy words directly from the \emph{value} portions of records in the content plan (i.e., $\{z_{k}\}_{k=1}^{|z|}$).
We experimented with joint \cite{P16-1154} and conditional
copy methods~\cite{P16-1014}.
Specifically, we introduce a variable $u_t \in \{0,1\}$ for each time step to indicate whether the predicted token $y_t$ is copied ($u_t = 1$) or not ($u_t = 0$).
The probability of generating $y_t$ is computed by:
\begin{align}
p(y_t|y_{<t},z,r) = \sum_{u_t \in \{ 0, 1 \}} p(y_t , u_t | y_{<t},z,r) \nonumber
\end{align}
where $u_t$ is marginalized out.

\paragraph{Joint Copy}
The probability of copying from record values and generating from
the vocabulary is globally normalized:
\begin{align}
&p(y_t , u_t | y_{<t},z,r) \propto \nonumber \\
&\begin{cases}
\sum_{y_t \leftarrow z_{k}} \exp (\mathbf{d}_t^\intercal \mathbf{W}_b \mathbf{e}_k) & u_t=1 \\
\exp{( \mathbf{W}_y \mathbf{d}_t^{att} + \mathbf{b}_y )} & u_t=0
\end{cases} \nonumber
\end{align}
where $y_t \leftarrow z_{k}$ indicates that~$y_t$ can be copied from $z_k$, $\mathbf{W}_b$ is shared as in Equation~\eqref{eq:beta:attention}, and $\mathbf{W}_y, \mathbf{b}_y$ are shared as in Equation~\eqref{eq:prb:gen}.

\paragraph{Conditional Copy}
The variable $u_t$ is first computed as a switch gate, and then is used to obtain the output probability:
\begin{align}
&p(u_t = 1 | y_{<t},z,r) = \sigmoid( \mathbf{w}_u \cdot \mathbf{d}_t + b_u ) \nonumber \\
&p(y_t , u_t | y_{<t},z,r) = \nonumber \\
&\begin{cases}
p(u_t|y_{<t},z,r) \sum_{y_t \leftarrow z_{k}} \beta_{t,k} & u_t=1 \\
p(u_t|y_{<t},z,r) p_{gen} (y_t | y_{<t},z,r) & u_t=0
\end{cases} \nonumber
\end{align}
where $\beta_{t,k}$ and $p_{gen} (y_t | y_{<t},z,r)$ are computed as
in~\Crefrange{eq:beta:attention}{eq:prb:gen}, and $\mathbf{w}_u \in
\mathbb{R}^{n}, b_u \in \mathbb{R}$ are parameters.  Following
\citeauthor{P16-1014} \shortcite{P16-1014} and
Wiseman et al.~\shortcite{wiseman2017challenges},
if $y_t$ appears in the content plan during training, we assume that
$y_t$~is copied (i.e., $u_t = 1$).\footnote{Following previous work
  (Gulcehre et al.~\citeyear{P16-1014}; Wiseman et
  al.~\citeyear{wiseman2017challenges}) we learn whether $y_t$ can be
  copied from candidate~$z_{k}$ by applying supervision during
  training. Specifically, we retain~$z_k$ when the record
  entity and its value occur in same sentence in~$y$.}

\subsection{Training and Inference}
Our model is trained to maximize the log-likelihood of the
gold\footnote{Strictly speaking, the content plan is not gold since it
  was not created by an expert but is the output of a fairly accurate
  IE system.}  content plan given table records~$r$ \emph{and} the
gold output text given the content plan and table records:
\begin{equation}
\max \sum_{(r, z, y) \in \mathcal{D} }{ \log{p \left( z | r \right)} + \log{p \left( y | r, z \right)} } \nonumber
\end{equation}
where $\mathcal{D}$ represents training examples (input records,
plans, and game summaries). During inference, the output for input $r$
is predicted by:
\begin{align}
\hat{z} &= \argmax_{z'} p(z' | r) \nonumber \\
\hat{y} &= \argmax_{y'} p(y' | r,\hat{z}) \nonumber
\end{align}
where $z'$ and $y'$ represent content plan and output text candidates,
respectively. For each stage, we utilize beam search to approximately
obtain the best results.

\section{Experimental Setup}
\label{sec:experimental-setup}

\paragraph{Data}
We trained and evaluated our model on \textsc{RotoWire} Wiseman et
al.~\shortcite{wiseman2017challenges}, a dataset of basketball game
summaries, paired with corresponding box- and line-score tables. The
summaries are professionally written, relatively well structured and
long (337~words on average). The number of record types is~39, the
average number of records is~628, the vocabulary size is 11.3K~words
and token count is 1.6M. The dataset is ideally suited for
document-scale generation.  We followed the data partitions introduced
in Wiseman et al.~(\citeyear{wiseman2017challenges}): we trained on
3,398 summaries, tested on 728, and used 727~for validation.

\paragraph{Content Plan Extraction}

We extracted content plans from the \textsc{RotoWire} game summaries
following an information extraction (IE) approach. Specifically, we
used the IE system introduced in Wiseman et
al.~\shortcite{wiseman2017challenges} which identifies candidate
entity (i.e.,~player, team, and city) and value (i.e.,~number or
string) pairs that appear in the text, and then predicts the type (aka
relation) of each candidate pair. For instance, in the document in
Table~\ref{fig:example}, the IE system might identify the pair
``\lform{\small Jeff Teague, 20}'' and then predict that that their
relation is ``\lform{\small PTS}'', extracting the record
(\lform{\small Jeff Teague, 20, PTS}). Wiseman et
al.~\shortcite{wiseman2017challenges} train an IE system on
\textsc{RotoWire} by determining word spans which could represent
entities (i.e.,~by matching them against players, teams or cities in
the database) and numbers. They then consider each entity-number pair
in the same sentence, and if there is a record in the database with
matching entities and values, the pair is assigned the corresponding
record type or otherwise given the label ``none'' to indicate
unrelated pairs.

We adopted their IE system architecture which predicts relations by
ensembling 3 convolutional models and 3 bidirectional LSTM models. We
trained this system on the training portion of the \textsc{RotoWire}
corpus.\footnote{A bug in the code of Wiseman et
  al.~\shortcite{wiseman2017challenges} excluded number words
  from the output summary. We corrected the bug and this resulted in
  greater recall for the relations extracted from the summaries. See
  the supplementary material for more details.}  On held-out data it
achieved 94\% accuracy, and recalled approximately 80\% of the
relations licensed by the records. Given the output of the IE system,
a content plan simply consists of (entity, value, record type, h/v)
tuples in their order of appearance in a game summary (the content
plan for the summary in Table~\ref{fig:example} is shown in
Table~\ref{tbl:example-content-plan}).  Player names are pre-processed
to indicate the individual's first name and surname (see \lform{\small
  Isaiah} and \lform{\small Thomas} in
Table~\ref{tbl:example-content-plan}); team records are also
pre-processed to indicate the name of team's city and the team itself
(see \lform{\small Boston} and \lform{\small Celtics} in
Table~\ref{tbl:example-content-plan}).

\paragraph{Training Configuration}
We validated model hyperparameters on the development set. %
We did not tune the dimensions of word embeddings and LSTM hidden
layers; we used the same value of ~$600$ reported in Wiseman et
al.~\shortcite{wiseman2017challenges}.  We used one-layer pointer
networks during content planning, and two-layer LSTMs during text
generation.  Input feeding (Luong et al.~\citeyear{D15-1166}) was
employed for the text decoder.  We applied dropout~(Zaremba et
al.~\citeyear{zaremba2014recurrent}) at a rate of~$0.3$. Models
were trained for $25$~epochs with the Adagrad optimizer~(Duchi et
al.~\citeyear{duchi2011adaptive}); the initial learning rate
was~$0.15$, learning rate decay was selected from $\{0.5, 0.97\}$, and
batch size was~5. For text decoding, we made use of
BPTT~\cite{mikolov2010recurrent} and set the truncation size to~$100$.
We set the beam size to~$5$ during inference.  All models are
implemented in OpenNMT-py \cite{klein2017opennmt}.

\section{Results}
\label{sec:results}

\paragraph{Automatic Evaluation}
We evaluated model output using the metrics defined in Wiseman et
al. \shortcite{wiseman2017challenges}.  The idea is to employ a fairly
accurate IE system (see the description in
Section~\ref{sec:experimental-setup}) on the gold and automatic
summaries and compare whether the identified relations align or
diverge.

Let $\hat{y}$~be the gold output, and $y$~the system output.
\emph{Content selection} (CS) measures how well (in terms of precision
and recall) the records extracted from $y$~match those found
in~$\hat{y}$.  \emph{Relation generation} (RG) measures the factuality
of the generation system as the proportion of records extracted
from~$y$ which are also found in $r$ (in terms of precision and number
of unique relations). \emph{Content ordering} (CO) measures how well
the system orders the records it has chosen and is computed as the
normalized Damerau-Levenshtein Distance between the sequence of
records extracted from $y$~and~$\hat{y}$. In addition to these
metrics, we report BLEU \cite{papineni-EtAl:2002:ACL}, with 
human-written game summaries as reference.

Our results on the development set are summarized in
Table~\ref{generation-from-plan-dev}. We compare our Neural Content
Planning model (NCP for short) against the two encoder-decoder (ED)
models presented in Wiseman et al. \shortcite{wiseman2017challenges}
with joint copy (JC) and conditional copy (CC), respectively. In
addition to our own re-implementation of these models, we include the
best scores reported in Wiseman et
al.~\shortcite{wiseman2017challenges} which were obtained with an
encoder-decoder model enhanced with conditional copy.
Table~\ref{generation-from-plan-dev} also shows results when NCP uses
oracle content plans (OR) as input. In addition, we report the
performance of a template-based generator (Wiseman et
al. \citeyear{wiseman2017challenges}) which creates a document
consisting of eight template sentences: an introductory sentence (who
won/lost), six player-specific sentences (based on the six
highest-scoring players in the game), and a conclusion sentence.

\begin{table}[t]
\small
\centering
\begin{tabular}{@{~}l@{~}|@{~}c@{~~~}c@{~}|c@{~~~}c|c|@{~}c@{~} }
 \thickhline
 \multirow{2}{*}{Model} &\multicolumn{2}{c|}{RG} &\multicolumn{2}{c|}{CS} & CO & \multirow{2}{*}{BLEU}\\

 &\# & P\% & P\% & R\% & DLD\% & \\ \thickhline
TEMPL &54.29 &99.92 &26.61 &59.16 &14.42 &8.51  \\
WS-2017 & 23.95 & 75.10 & 28.11 & 35.86 & 15.33 & 14.57 \\
ED+JC & 22.98 & 76.07 & 27.70 & 33.29 & 14.36 & 13.22  \\ %
ED+CC & 21.94 & 75.08 &27.96 & 32.71 & 15.03 & 13.31 \\ %
NCP+JC  & 33.37 & 87.40 & 32.20 & 48.56 & 17.98 & 14.92 \\ %
NCP+CC &{33.88} & {87.51} & {33.52} & {51.21}
 &{18.57} & {16.19}\\ %
NCP+OR & 21.59 & 89.21 & 88.52 & 85.84 & 78.51 & 24.11\\
\thickhline
\end{tabular}
\caption{\label{generation-from-plan-dev} Automatic evaluation
  on  \textsc{RotoWire} development set using 
  relation generation (RG)  count (\#) and precision (P\%), content
  selection (CS) precision (P\%) and recall (R\%), content ordering (CO) in normalized
  Damerau-Levenshtein distance (DLD\%), and BLEU.}%
\end{table}

\begin{table}[t]
\small
\centering
\begin{tabular}{ @{~}l@{~}|@{~}c@{~~}c|c@{~~}c|c|c@{~} } 
 \thickhline
\multirow{2}{*}{Model} &\multicolumn{2}{c|}{RG} &\multicolumn{2}{c|}{CS} & CO & \multirow{2}{*}{BLEU}\\ 

 &\# & P\% & P\% & R\% & DLD\% & \\ 
\thickhline
ED+CC & 21.94 & 75.08 &27.96 & 32.71 & 15.03 & 13.31 \\ %
CS+CC &24.93 &80.55 &28.63 &35.23 &15.12 &13.52 \\ %
CP+CC &33.73 &84.85 &29.57 &44.72 &15.84 &14.45 \\ %
NCP+CC &{33.88} & {87.51} & {33.52} & {51.21}
 &{18.57} & {16.19}\\ %
NCP & 34.46 & ---& 38.00 & 53.72 & 20.27& --- \\ 
\thickhline
\end{tabular}
\caption{\label{tbl:ablation-content-selection-planning}Ablation
  results on  \textsc{RotoWire} development set using 
  relation generation (RG)  count (\#) and precision (P\%), content
  selection (CS) precision (P\%) and recall (R\%), content ordering
  (CO) in normalized Damerau-Levenshtein distance (DLD\%),
  and BLEU.}
\end{table}

As can be seen, NCP improves upon vanilla encoder-decoder models
(ED+JC, ED+CC), irrespective of the copy mechanism being employed. In
fact, NCP achieves comparable scores with either joint or conditional
copy mechanism which indicates that it is the content planner which
brings performance improvements. Overall, NCP+CC achieves best content
selection and content ordering scores in terms of BLEU. Compared to
the best reported system in Wiseman et
al. \shortcite{wiseman2017challenges}, we achieve an absolute
improvement of approximately 12\% in terms of relation generation;
content selection precision also improves by~5\% and recall by~15\%,
content ordering increases by 3\%, and BLEU by 1.5~points. The results
of the oracle system (NCP+OR) show that content selection and ordering
do indeed correlate with the quality of the content plan and that any
improvements in our planning component would result in better
output. As far as the template-based system is concerned, we observe
that it obtains low BLEU and CS precision but scores high on CS recall
and RG metrics.  This is not surprising as the template system is
provided with domain knowledge which our model does not have, and thus
represents an upper-bound on content selection and relation
generation. We also measured the degree to which the game summaries
generated by our model contain redundant information as the proportion
of non-duplicate records extracted from the summary by the IE system.
84.5\% of the records in NCP+CC are non-duplicates compared to Wiseman
et al.  \shortcite{wiseman2017challenges} who obtain 72.9\% showing
that our model is less repetitive.

\begin{table}[t]
\small
\centering
\begin{tabular}{@{~}l@{~}|@{~}c@{~~~}c@{~}|c@{~~~}c|c|@{~}c@{~} } 
 \thickhline
 \multirow{2}{*}{Model} &\multicolumn{2}{c|}{RG} &\multicolumn{2}{c|}{CS} & CO & \multirow{2}{*}{BLEU}\\ 
 &\# & P\% & P\% & R\% & DLD\% & \\ \thickhline
TEMPL &\textbf{54.23} &\textbf{99.94} &26.99 &\textbf{58.16} &14.92 &\hspace*{1ex}8.46  \\
WS-2017 & 23.72 & 74.80 & 29.49 & 36.18 & 15.42 & 14.19 \\
NCP+JC & 34.09 & 87.19 & 32.02 & 47.29 & 17.15 & 14.89 \\ %
NCP+CC &{34.28} & {87.47} & \textbf{34.18} & {51.22}
 &\textbf{18.58} & \textbf{16.50}\\ %
\thickhline
\end{tabular}
\caption{\label{generation-from-plan-test} Automatic evaluation
  on  \textsc{RotoWire} test set using   relation generation (RG)
  count (\#) and precision (P\%), content selection (CS) precision
  (R\%) and recall (R\%), content ordering (CO) in normalized
  Damerau-Levenshtein distance (DLD\%), and BLEU.} 
\end{table}

We further conducted an ablation study with the conditional copy
variant of our model (NCP+CC) to establish whether improvements
are due to better content selection (CS) and/or content planning (CP).
We see in Table~\ref{tbl:ablation-content-selection-planning} that
content selection and planning individually contribute to performance
improvements over the baseline (ED+CC), and accuracy further increases
when both components are taken into account.  In addition we evaluated
these components on their own (independently of text generation) by
comparing the output of the planner (see $p(z|r)$ block in
Figure~\ref{fig:overall-method}) against gold content plans obtained
using the IE system (see row NCP in
Table~\ref{tbl:ablation-content-selection-planning}.  Compared to the
full system (NCP+CC), content selection precision and recall are
higher (by~4.5\% and 2\%, respectively) as well as content ordering
(by~1.8\%). In another study, we used the CS and CO metrics to measure
how well the generated text follows the content plan produced by the
planner (instead of arbitrarily adding or removing information). We
found out that NCP+CC generates game summaries which follow the
content plan closely: CS precision is higher than 85\%, CS recall is
higher than 93\%, and CO higher than 84\%. This reinforces our claim
that higher accuracies in the content selection and planning phases
will result in further improvements in text generation.

The test set results in Table~\ref{generation-from-plan-test} follow a
pattern similar to the development set.  NCP achieves higher accuracy
in all metrics including relation generation, content selection,
content ordering, and BLEU compared to Wiseman et
al.~\shortcite{wiseman2017challenges}. We provide examples of system
output in Table~\ref{tab:output} and the supplementary material.

\begin{table}[t]
\scriptsize
\centering
\begin{tabular}{@{~}p{8.3cm}@{~}} \thickhline
{\color{blue}The Golden State Warriors (10--2) defeated the
  Boston Celtics (6--6) 104--88. Klay Thompson scored 28 points
  (12--21 FG, 3--6 3PT, 1--1 FT) to go with 4 rebounds. Kevin Durant
  scored 23 points (10--13 FG, 1--2 3PT, 2--4 FT) to go with 10
  rebounds. Isaiah Thomas scored 18 points (4--12 FG, 1--6 3PT, 9--9
  FT) to go with 2 rebounds. Avery Bradley scored 17 points (7--15 FG,
  2--4 3PT, 1--2 FT) to go with 10 rebounds. Stephen Curry  scored 16
  points (7--20 FG, 2--10 3PT, 0--0 FT) to go with 3  rebounds. Terry
  Rozier scored 11 points (3--5 FG, 2--3 3PT, 3--4 FT)  to go with 7
  rebounds.} The Golden State Warriors' next game will be at home
against the Dallas Mavericks, while the Boston Celtics will travel to
play the Bulls. \\ \hline 
{\color{blue}The Golden State Warriors defeated the Boston Celtics 104--88 }at TD Garden on Friday. {\color{blue}The Warriors (10--2)}
came into this game winners of five of their last six games, but the
{\color{red}Warriors (6--6)} were able to pull away in the second half. {\color{blue}Klay Thompson led the way for the Warriors with 28 points on 12--of--21 shooting, while Kevin Durant added 23 points, 10 rebounds, seven assists and two steals. Stephen Curry added 16 points and eight assists, while Draymond Green rounded out the box score with 11 points, eight rebounds and eight assists. For the Celtics, it was Isaiah Thomas who shot just 4--of--12 from the field and finished with 18 points. Avery Bradley added 17 points and 10 rebounds}, while the rest of the Celtics {\color{red}combined to score just seven points}. Boston will look to get back on track as they play host to the 76ers on Friday. \\
\thickhline
\end{tabular}
\caption{\label{tab:output} Example output from TEMPL (top)
  and NPC+CC (bottom). Text that accurately reflects
  a record in the associated box or line score is in {\color{blue}blue}, erroneous text
  is in {\color{red}red.}}
\end{table}

\paragraph{Human-Based Evaluation}
We conducted two human evaluation experiments using the Amazon
Mechanical Turk (AMT) crowdsourcing platform. The first study assessed
relation generation by examining whether improvements in relation
generation attested by automatic evaluation metrics are indeed
corroborated by human judgments.  We compared our best performing
model (NCP+CC), with gold reference summaries, a template system and
the best model of Wiseman et
al. \shortcite{wiseman2017challenges}. AMT workers were presented with
a specific NBA game's box score and line score, and four (randomly
selected) sentences from the summary. They were asked to identify
supporting and contradicting facts mentioned in each sentence. We
randomly selected 30~games from the test set. Each sentence was rated
by three workers.

The left two columns in Table~\ref{tbl:human-eval} contain the average
number of supporting and contradicting facts per sentence as
determined by the crowdworkers, for each model.  The template-based
system has the highest number of supporting facts, even compared to
the human gold standard. TEMPL does not perform any content selection,
it includes a large number of facts from the database and since it
does not perform any generation either, it exhibits a few
contradictions.  Compared to WS-2017 and the Gold summaries, NCP+CC
displays a larger number of supporting facts. All models are
significantly\footnote{All significance differences reported
  throughout this paper are with a level less than 0.05.}  different
in the number of supporting facts (\#Supp) from TEMPL (using a one-way
ANOVA with post-hoc Tukey HSD tests). NCP+CC is significantly
different from WS-2017 and Gold.  With respect to contradicting facts
(\#Cont), Gold and TEMPL are not significantly different from each
other but are significantly different from the neural systems
(WS-2017, NCP+CC). %

In the second experiment, we assessed the generation quality of our
model.  We elicited judgments for the same 30~games used in the first
study. For each game, participants were asked to compare a
human-written summary, NCP with conditional copy (NCP+CC), Wiseman et
al.'s \shortcite{wiseman2017challenges} best model, and the template
system.  
Our study used \textit{Best-Worst Scaling} (BWS;
\citeauthor{louviere2015best} \citeyear{louviere2015best}), a technique shown to be
less labor-intensive and providing more reliable results as compared to 
rating scales \cite{kiritchenko2017best}. We arranged every 4-tuple of competing
summaries into 6~pairs. Every pair was shown to three crowdworkers,
who were asked to choose which summary was {\it best} and which was {\it worst}
according to three criteria: \emph{Grammaticality}
(is the summary fluent and grammatical?), \emph{Coherence} (is the
summary easy to read? does it follow a natural ordering of facts?),
and \emph{Conciseness} (does the summary avoid redundant information
and repetitions?). The score of a system for each criterion is computed as the difference between the percentage
of times the system was selected as the best and the percentage of times it was
selected as the worst \cite{orme2009maxdiff}. The scores range from $-100$
(absolutely worst) to $+100$ (absolutely best).

\begin{table}[t]
\small
\centering
\begin{tabular}{@{~}lccrrr@{~}}
\thickhline 
 & \#Support & \#Contra & Gram & Cohere & Concise \\
\thickhline 
Gold    & 2.98 & 0.28 & 11.78 & 16.00& 13.78 \\
TEMPL   & {6.98} & {0.21} & -0.89 &-4.89 & 1.33\\
WS-2017 & 3.19 & 1.09 & -4.22 &-4.89 &-6.44 \\
NCP+CC  & 4.90 & 0.90 & -2.44 & -2.44&-3.55\\
\thickhline 
\end{tabular} 
\caption{Average number of supporting (\#Support) and contradicting
  (\#Contra) facts
  in game summaries and   \textit{best-worst scaling}
  evaluation (higher is better) for grammaticality (Gram), Coherence (Cohere), and
  Conciseness (Concise).}
\label{tbl:human-eval}
\end{table}

The results of the second study are summarized in
Table~\ref{tbl:human-eval}.  Gold summaries were perceived as
significantly better compared to the automatic systems across all
criteria (again using a one-way ANOVA with post-hoc Tukey HSD
tests). NCP+CC was perceived as significantly more grammatical than
WS-2017 but not compared to TEMPL which does not suffer from
fluency errors since it does not perform any generation. NCP+CC was
perceived as significantly more coherent than TEMPL and WS-2017. The
template fairs poorly on coherence, its output is stilted and exhibits
no variability (see top block in Table~\ref{tab:output}). With regard
to conciseness, the neural systems are significantly worse than TEMPL,
while NCP+CC is significantly better than WS-2017. By design the
template cannot repeat information since there is no redundancy in the
sentences chosen to verbalize the summary.

Taken together, our results show that content planning improves
data-to-text generation across metrics and systems. We find that
NCP+CC overall performs best, however there is a significant gap
between automatically generated summaries and human-authored ones.

\section{Conclusions}
\label{sec:conclusions}

We presented a data-to-text generation model which is enhanced with
content selection and planning modules. Experimental results (based on
automatic metrics and judgment elicitation studies) demonstrate that
generation quality improves both in terms of the number of relevant
facts contained in the output text, and the order according to which
these are presented. Positive side-effects of content planning are
additional improvements in the grammaticality, and conciseness of the
generated text.  In the future, we would like to learn more
detail-oriented plans involving inference over multiple facts and
entities. We would also like to verify our approach across domains and
languages.

\fontsize{9.0pt}{10.0pt} \selectfont
\bibliography{aaai2019}
\bibliographystyle{aaai}
\section*{Comparison with the Results in Wiseman~et~al.'s~\shortcite{wiseman2017challenges} Webpage}
\label{sec:supplemental}
There was a bug in the dataset creation of Wiseman et
al.~\shortcite{wiseman2017challenges} which they identified and
corrected. They also posted updated scores on their
webpage.\footnote{https://github.com/harvardnlp/data2text} We have
used this corrected dataset in our experiments. We then discovered a
bug in their code which computes the automatic metrics. The scores
reported in this paper are using the corrected automatic metrics. To
make the scores on our paper comparable to the numbers published on
the webpage of Wiseman et al.~\shortcite{wiseman2017challenges}, 
we recompute here our
scores with older IE metrics (without the bug fix) in
Table~\ref{generation-from-plan-dev-bug} (development set) and
Table~\ref{generation-from-plan-test-bug} (test set).

\section*{Qualitative Examples}
Table~\ref{tbl:examples} shows two sample documents generated using
the template system, Wiseman et al.~\shortcite{wiseman2017challenges} (WS-2017) and our
neural content planning model with conditional copy (NCP+CC). The text
is highlighted in blue if it agrees with respective box/line scores
and red if the text contradicts box/line scores. We also use the
orange color to highlight repetitions.

The template documents are gold standard in relation generation
accuracy and they appear all in blue. The documents of
Wiseman et al.~\shortcite{wiseman2017challenges} show instances of contradictions and
tend to be verbose containing duplicate text too. In contrast, our
neural content planning model generates more factual text with fewer
contradictions to box/line scores and less duplicate information.

\begin{table}[h]
\small
\centering
\begin{tabular}{@{~}l@{~}|@{~}c@{~~~}c@{~}|c@{~~~}c|c|@{~}c@{~} }
 \thickhline
 \multirow{2}{*}{Model} &\multicolumn{2}{c|}{RG} &\multicolumn{2}{c|}{CS} & CO & \multirow{2}{*}{BLEU}\\

 &\# & P\% & P\% & R\% & DLD\% & \\ \thickhline
TEMPL &54.22 &99.92 &23.41 &72.62 &11.30 &8.51  \\
WS-2017 & 16.93 & 75.74 & 31.2 & 38.94 & 14.98 & 14.57 \\
ED+JC & 16.65 & 77.82 & 31.79 & 37.82 & 14.50 & 13.22  \\ %
ED+CC & 16.5 & 74.7 &29.87 & 37.00 & 14.10 & 13.31 \\ %
NCP+JC  & 24.74 & 88.66 & 34.80 & 53.67 & 16.34 & 14.92 \\ %
NCP+CC &24.95 & 88.08 & 35.50 & 55.36
 &17.23 & 16.19\\
NCP+OR & 16.11 & 89.35 & 86.39 & 87.12 & 52.25 & 24.11\\
\thickhline
\end{tabular}
\caption{\label{generation-from-plan-dev-bug} Automatic system evaluation
  on the \textsc{RotoWire} development set using automatic evaluation
  metrics;  relation generation (RG)  count and precision, content selection (CS) precision and recall, content ordering (CO) in normalized
  Damerau-Levenshtein distance, and BLEU. TEMPL is template system, WS-2017 is the best system of Wiseman et al.~\shortcite{wiseman2017challenges},
  ED+JC, ED+CC are encoder-decoder with joint copy and conditional copy, respectively, NCP+JC, NCP+CC are our Neural Content Planning models
  with joint copy and conditional copy, respectively. NCP+OR is the system with oracle content plan as input.}
\end{table}

\begin{table}[t]
\small
\centering
\begin{tabular}{@{~}l@{~}|@{~}c@{~~~}c@{~}|c@{~~~}c|c|@{~}c@{~} }
 \thickhline
 \multirow{2}{*}{Model} &\multicolumn{2}{c|}{RG} &\multicolumn{2}{c|}{CS} & CO & \multirow{2}{*}{BLEU}\\

 &\# & P\% & P\% & R\% & DLD\% & \\ \thickhline
TEMPL &\textbf{54.17} &\textbf{99.95} &23.74 &\textbf{72.36} &11.68 &8.46  \\
WS-2017 & 16.83 & 75.62 & 32.80 & 39.93 & 15.62 & 14.19 \\
NCP+JC  & 26.14 & 90.74 & 34.15 & 53.61 & 15.94 & 14.89 \\ %
NCP+CC &25.41 & 88.31 & \textbf{36.07} & 56.17
 &\textbf{16.94} & \textbf{16.50}\\
\thickhline
\end{tabular}
\caption{\label{generation-from-plan-test-bug} Automatic system evaluation
  on the \textsc{RotoWire} test set using automatic evaluation
  metrics;  relation generation (RG)  count and precision, content selection (CS) precision and recall, content ordering (CO) in normalized
  Damerau-Levenshtein distance, and BLEU. TEMPL is template system, WS-2017 is the best system of Wiseman et al.~\shortcite{wiseman2017challenges},
  NCP+JC, NCP+CC are our Neural Content Planning models
  with joint copy and conditional copy, respectively.}
\end{table}

\begin{table*}
\small
\centering
\begin{tabular}{|c|p{15.5cm}|}
\hline 
System & \multicolumn{1}{c|}{Summary} \\ 
\hline 
Template & {\color{blue} The Washington Wizards (8-13) defeated the Denver Nuggets (8-15) 92-85. Bradley Beal scored 26 points (8-15 FG, 4-7 3PT, 6-6 FT) to go with 3 rebounds. Nikola Jokic scored 17 points (6-10 FG, 0-0 3PT, 5-7 FT) to go with 11 rebounds. Markieff Morris scored 15 points (5-12 FG, 0-0 3PT, 5-5 FT) to go with 3 rebounds. John Wall scored 15 points (5-14 FG, 0-4 3PT, 5-6 FT) to go with 7 rebounds. Danilo Gallinari scored 14 points (3-11 FG, 1-8 3PT, 7-9 FT) to go with 4 rebounds. Jusuf Nurkic scored 13 points (6-6 FG, 0-0 3PT, 1-2 FT) to go with 7 rebounds.} The Washington Wizards' next game will be at home against the Dallas Mavericks, while the Denver Nuggets will travel to play the Bulls. \\ 
\hline 
WS-2017 & {\color{blue} The Washington Wizards (8-13) defeated the Denver Nuggets (8-15) 92-85} on Wednesday at the Verizon Center in Washington. {\color{red}The Wizards got off to a hot start in this one, out-scoring the Nuggets 29-18 in the first quarter alone. The Nuggets were able to out-score the Nuggets 28-12 in the third quarter}, while the Wizards were able to coast to a victory in front of their home crowd. {\color{red}The Nuggets were the superior shooters in this game, going 45 percent from the field and 32 percent from the three-point line, while the Wizards went just 41 percent from the floor and 32 percent from beyond the arc}. {\color{blue}The Nuggets were also able to shoot 41 percent from the field} {\color{red}and 32 percent from the three-point line }, while the {\color{blue}Nuggets went just 41 percent from the floor} and {\color{red}32 percent from beyond the arc}. {\color{red}The Nuggets were led by the duo of John Wall and Bradley Beal}. {\color{blue}Wall went 5-for-14 from the field and 0-}for{\color{red}-4 }from the three-point line to score a game-{\color{blue}high of 15 points, while also adding seven rebounds and five assists}. It was his second double-double in a row, as he's combined for 47 points and 12 rebounds over his last two games. The only other Nugget to reach double figures in points was Wilson Chandler, {\color{red}who finished with 11 points} and {\color{blue}eight rebounds}. It was his second double-double in a row, as he's combined for 47 points and 14 rebounds over that span. {\color{orange}The only other Nugget to reach double figures in points was Wilson Chandler, {\color{red}who finished with 11 points} and {\color{blue} eight rebounds}}. The Nuggets' next game will be on the road against the Detroit Pistons on Friday, while the Nuggets will be at home against the New York Knicks on Friday. \\ 
\hline 
NCP+CC & {\color{blue} The Washington Wizards defeated the visiting Denver Nuggets 92-85} at Verizon Center on Monday. {\color{blue} The Wizards (8-13)} came into this game winners of five of their last eight games, but the {\color{red} Wizards (8-15)} jumped out to a 10-point lead at the end of the first quarter. {\color{blue} Bradley Beal led the way for the Wizards with a game-high 26 points on 8-of-15 shooting from the field. John Wall shot 5-of-14 from the field on his way to 15 points, to go along with seven rebounds, five assists, three steals and one block. Jusuf Nurkic chipped in 13 points, seven rebounds and one assist. Jameer Nelson filled out the stat sheet with 10 points, eight assists, four rebounds and two steals in 36 minutes}. As a team, it was a forgettable shooting night for the Nuggets, as {\color{red} the team shot just 46 percent from the field}. Next up, the Nuggets play the second game of a back-to-back when they host the Denver Nuggets on Wednesday, while the Wizards host the Portland Trail Blazers on Friday. \\ 
\hline 
\hline 
Template &{\color{blue} The Golden State Warriors (10-2) defeated the Boston Celtics (6-6) 104-88. Klay Thompson scored 28 points (12-21 FG, 3-6 3PT, 1-1 FT) to go with 4 rebounds. Kevin Durant scored 23 points (10-13 FG, 1-2 3PT, 2-4 FT) to go with 10 rebounds. Isaiah Thomas scored 18 points (4-12 FG, 1-6 3PT, 9-9 FT) to go with 2 rebounds. Avery Bradley scored 17 points (7-15 FG, 2-4 3PT, 1-2 FT) to go with 10 rebounds. Stephen Curry scored 16 points (7-20 FG, 2-10 3PT, 0-0 FT) to go with 3 rebounds. Terry Rozier scored 11 points (3-5 FG, 2-3 3PT, 3-4 FT) to go with 7 rebounds.} The Golden State Warriors' next game will be at home against the Dallas Mavericks, while the Boston Celtics will travel to play the Bulls. \\ 
\hline 
WS-2017 & {\color{blue}The Golden State Warriors defeated the Boston Celtics, 104-88}, at TD Garden on Wednesday. The {\color{blue}Warriors (10-2)} checked in to Saturday's contest with only two road wins in their last 11 games, but they were able to come away with a win against the {\color{blue}Celtics (6-6)} on Friday. The {\color{blue}Warriors (10-2)} were able to pull away in the second half, {\color{red}outscoring the Celtics (6-6) by a 31-9 margin }over the final 12 minutes. However, Golden State was able to pull away in the second half, {\color{red}outscoring the Celtics by a 31-9 margin} over the final 12 minutes. The {\color{red}Warriors were led by Kevin Durant}'s {\color{blue}23 points}, which he supplemented with {\color{red}seven} rebounds, {\color{blue}seven assists, two steals and a block}. Stephen Curry was next {\color{blue}with 16 points, eight assists, three rebounds and four steals}. {\color{blue}Klay Thompson was next with a 28-point}, {\color{red}10-rebound} double-double that also included {\color{blue}three assists}, {\color{red}two steals and a block}. {\color{blue}Draymond Green was next with 11 points, eight rebounds, eight assists and two blocks. {\color{orange}Draymond Green was next with 11 points, eight rebounds, eight assists and two blocks. Draymond Green supplied 11 points, eight rebounds, eight assists, two blocks and a steal}}. {\color{blue}David West paced the reserves with 4 points, two rebounds, a block and a block}. The Celtics were paced by {\color{blue}Thomas' 18 points, which he supplemented with four assists, two rebounds} and {\color{red}four steals}. {\color{blue}Avery Bradley posted a 17-point, 10-rebound double-double that also included two assists, two steals and a block. {\color{orange}Avery Bradley posted a 17-point, 10-rebound double-double that also included two assists, two steals and a block}}. {\color{red}Kelly Olynyk led the second unit with 11 points}, {\color{blue}three rebounds, two assists} and {\color{red}a pair of steals}. The Warriors head back home to face off with the Detroit Pistons on Friday night, while the Celtics remain home and await the Toronto Raptors for a Wednesday night showdown. \\ 
\hline 
NCP+CC & {\color{blue}The Golden State Warriors defeated the Boston Celtics 104-88 }at TD Garden on Friday. {\color{blue}The Warriors (10-2)} came into this game winners of five of their last six games, but the {\color{red}Warriors (6-6)} were able to pull away in the second half. {\color{blue}Klay Thompson led the way for the Warriors with 28 points on 12-of-21 shooting, while Kevin Durant added 23 points, 10 rebounds, seven assists and two steals. Stephen Curry added 16 points and eight assists, while Draymond Green rounded out the box score with 11 points, eight rebounds and eight assists. For the Celtics, it was Isaiah Thomas who shot just 4-of-12 from the field and finished with 18 points. Avery Bradley added 17 points and 10 rebounds}, while the rest of the Celtics {\color{red}combined to score just seven points}. Boston will look to get back on track as they play host to the 76ers on Friday. \\ 
\hline 
\end{tabular} 
\caption{Example documents from the template-based system, WS-2017,
  the best system of Wiseman et al.~\shortcite{wiseman2017challenges},  and 
  our Neural Content Planning model with conditional copy (NCP+CC). Text that accurately reflects
  a record in the associated box or line score is recorded in {\color{blue}blue}, erroneous text
  is marked in {\color{red}red}, duplicate text is marked in {\color{orange}orange}.}
\label{tbl:examples}
\end{table*}

\end{document}